\documentclass[letterpaper]{article} 
\usepackage{aaai25}  
\usepackage{times}  
\usepackage{helvet}  
\usepackage{courier}  
\usepackage[hyphens]{url}  
\usepackage{graphicx} 
\usepackage{natbib}  
\usepackage{caption} 
\usepackage{algorithm}
\usepackage{algpseudocode}
\usepackage{pgfplots}
\usepackage{amsmath} 
\usepackage{booktabs}

\usepackage{tikz}
\usetikzlibrary{shadows, shapes, arrows.meta, positioning}

\usepackage{newfloat}
\usepackage{listings}
\DeclareCaptionStyle{ruled}{labelfont=normalfont,labelsep=colon,strut=off} 
\floatstyle{ruled}
\newfloat{listing}{tb}{lst}{}
\floatname{listing}{Listing}
\title{Seq2Seq Model-Based Chatbot with LSTM and Attention Mechanism for Enhanced User Interaction}
\author {
   Lamya Benaddi\textsuperscript{\rm 1},
    Charaf Ouaddi\textsuperscript{\rm 1},
    Adnane Souha\textsuperscript{\rm 1},
    Abdeslam Jakimi\textsuperscript{\rm 1},
    Mohamed Rahouti\textsuperscript{\rm 2},
    Mohammed Aledhari\textsuperscript{\rm 3},
    Diogo Oliveira\textsuperscript{\rm 4},
    Brahim Ouchao\textsuperscript{\rm 1}
}
\affiliations {
    \textsuperscript{\rm 1}GL-ISI Team, Department of Computer Sciences, Faculty of Science and Technology Errachidia, Moulay Ismail University, Morocco\\
    \textsuperscript{\rm 2}CIS Dept. Fordham Univ. New York, NY USA \\

     \textsuperscript{\rm 3}Dept. of Data Science, Univ. of North texas Denton, TX, USA \\

      \textsuperscript{\rm 4}Point Park Univ. Pittsburgh, PA USA\\
   {l.benaddiedu.umi.ac.ma, c.ouaddiedu.umi.ac.ma, ad.souha}@edu.umi.ac.ma, ajakimi@yahoo.fr, mrahouti@fordham.edu, Mohammed.Aledhari@unt.edu, diogo.oliveira@pointpark.edu, ouchaob@gmail.com
}

\usepackage{bibentry}

\begin{document}

\maketitle

\begin{abstract}
A chatbot is an intelligent software application that automates conversations and engages users in natural language through messaging platforms. Leveraging artificial intelligence (AI), chatbots serve various functions, including customer service, information gathering, and casual conversation. Existing virtual assistant chatbots, such as ChatGPT and Gemini, demonstrate the potential of AI in Natural Language Processing (NLP). However, many current solutions rely on predefined APIs, which can result in vendor lock-in and high costs. To address these challenges, this work proposes a chatbot developed using a Sequence-to-Sequence (Seq2Seq) model with an encoder-decoder architecture that incorporates attention mechanisms and Long Short-Term Memory (LSTM) cells. By avoiding predefined APIs, this approach ensures flexibility and cost-effectiveness. The chatbot is trained, validated, and tested on a dataset specifically curated for the tourism sector in Draa-Tafilalet, Morocco. Key evaluation findings indicate that the proposed Seq2Seq model-based chatbot achieved high accuracies: approximately 99.58\% in training, 98.03\% in validation, and 94.12\% in testing. These results demonstrate the chatbot's effectiveness in providing relevant and coherent responses within the tourism domain, highlighting the potential of specialized AI applications to enhance user experience and satisfaction in niche markets.
\end{abstract}

%

\section{Introduction}

A chatbot is an advanced software application designed to facilitate automated conversations and engage users in natural language through messaging platforms \cite{1}. Leveraging artificial intelligence (AI), chatbots serve various functions, including customer service, information retrieval, and casual conversation. Contemporary virtual assistant chatbots, such as ChatGPT, Gemini, and others, exemplify the application of AI in this domain \cite{casheekar2024contemporary}.

The development of chatbots utilizing AI techniques represents a significant advancement in Natural Language Processing (NLP). Numerous studies employ deep learning and NLP methodologies to construct sophisticated chatbot systems. Additionally, researchers often utilize APIs provided by intent recognition services to enhance chatbot functionality \cite{2}. However, these APIs come with certain limitations, such as potential dependency on specific NLP service providers and associated high costs.

The proposed work addresses critical gaps in current state-of-the-art chatbots by developing a solution specifically tailored for the tourism sector in Draa-Tafilalet, Morocco \cite{DraaTafilalet}. While existing chatbots, such as ChatGPT and other general-purpose conversational agents, have made significant advancements in NLP and user interaction \cite{adamopoulou2020chatbots, limna2023use}, they often lack the domain-specific knowledge and cultural context required to provide accurate and relevant information for niche sectors \cite{zhai2022systematic}. This work employs a Sequence-to-Sequence (Seq2Seq) model with Long
Short-Term Memory (LSTM) and attention mechanisms to achieve a high level of contextual understanding and response generation, crucial for engaging users with specific interests in tourism.

By constructing the chatbot without relying on predefined APIs, this approach avoids potential lock-in and the high costs associated with commercial NLP services, offering a more flexible and cost-effective solution. The focus on domain-specific expertise, combined with advanced deep learning techniques, positions the proposed chatbot as a superior alternative to general-purpose bots, enhancing user experience and satisfaction in the tourism domain. To overcome the aforementioned limitations, the proposed methodology further leverages a Seq2Seq model with an encoder-decoder architecture, utilizing attention mechanisms and LSTM cells \cite{3}. A curated dataset of question/answer pairs specific to the tourism sector is used to train, validate, and test the model, ensuring robust performance and applicability.

The key contributions of this work are outlined as follows:
\begin{itemize}
    \item Develop a Seq2Seq model-based chatbot using LSTM cells and an attention mechanism to improve interaction quality.
    \item Curate a diverse dataset specific to the tourism sector, ensuring robust training, validation, and testing processes.
    \item Achieve high training, validation, and test accuracies, demonstrating the effectiveness of the chosen architecture and techniques.
    \item Design a chatbot capable of providing informative and engaging interactions for users in the tourism domain, showcasing real-world applicability.
\end{itemize}

The rest of this paper is organized as follows. We first provide a background of the chatbots and related work summary. After that, we present an overview of our proposed methodology. Next, We discuss our experiments and key evaluation findings. Finally, we conclude this work.

\section{Research Background and State-of-the-Art} \label{sec:related}

\begin{figure}[t]
\centering
\includegraphics[width=1\linewidth]{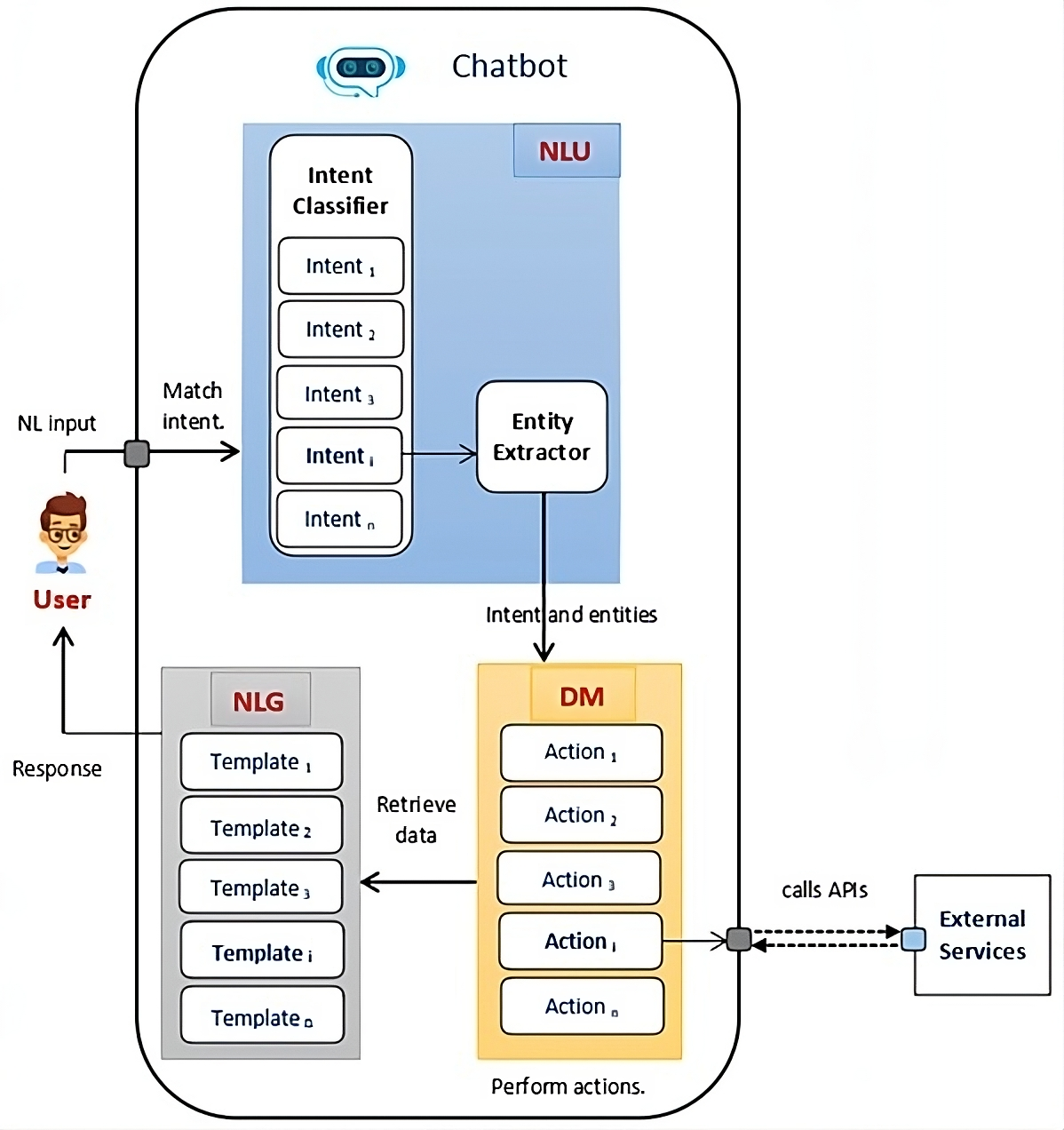}
\caption{ The conceptual architecture of chatbots \cite{13}.}
\label{fig:rule-based-chatbot}
\end{figure}

Chatbots, also known as conversational agents or virtual assistants \cite{4}, have emerged as significant and powerful tools across various domains such as commerce, education, and tourism, fundamentally transforming human-technology interaction. A chatbot is a computer program designed to conduct conversations with human users via textual or auditory interfaces. The concept of chatbots dates back to the 1960s, with early prototypes like ELIZA \cite{5}, but it was not until recent advancements in deep learning and NLP that chatbots achieved remarkable maturity for conducting human-like dialogues.

Chatbots can be broadly classified into two categories: \textbf{rule-based chatbots} and \textbf{AI-based chatbots} \cite{6}. Rule-based chatbots operate on predefined rules and patterns programmed by developers. These chatbots follow a structured approach, relying on pre-defined decision trees or if-else conditions to generate responses. In contrast, AI-based chatbots employ deep learning techniques, such as Recurrent Neural Networks (RNNs) with LSTM \cite{7} or transformer architectures, and NLP to dynamically understand and generate user responses.

Moreover, several platforms and frameworks \cite{8,9} have emerged to facilitate the development of chatbots, each offering unique technical and managerial features. These platforms provide developers with tools for building, training, and deploying chatbots across various applications like websites, mobile apps, or social networks. Examples of such tools include Dialogflow, Rasa, Amazon Lex, IBM Watson, and others \cite{2}.

\subsection{Rule-Based Chatbots}

The architecture of rule-based chatbots \cite{10}, as illustrated in Fig. \ref{fig:rule-based-chatbot}, typically consists of three main components: Natural Language Understanding (NLU), Dialog Management (DM), and Natural Language Generation (NLG). This architecture is designed around a set of intents that users may want to accomplish. For instance, given a user input like ``I would like to buy three packs of pizza" (the first arrow in Fig. \ref{fig:rule-based-chatbot}), the chatbot tries to identify the corresponding intent. If the chatbot does not find any matching intent, some tools allow for a default fallback intent \cite{11}.

After matching an intent at the NLU level, the chatbot extracts the parameters from the utterance, such as the quantity of pizza packets (the second arrow). The chatbot, facilitated by its DM component, can execute diverse actions based on user intent, including interfacing with external services to retrieve data \cite{12}. Finally, the NLG component receives data from the DM and uses predefined templates to generate natural language responses and reply to the user.

\subsection{AI-Based Chatbots}

In contrast, AI-based chatbots adopt an end-to-end architecture where a single AI generative model handles the entire conversation pipeline, from understanding user inputs to generating responses. This architecture eliminates the need for explicit rule-based components, allowing the chatbot to learn and adapt to diverse conversational contexts.

AI-based chatbot architecture typically leverages advanced AI and NLP techniques. RNNs, a specialized class of artificial neural networks, are particularly well-suited for processing sequential data such as text and audio, making them ideal for chatbot applications. Unlike Feed-Forward Neural Networks (FFNN), which are constrained to fixed-length inputs, RNNs can handle sequences of varying lengths. By processing input sequences \(X\), RNNs generate a sequence of hidden states \(Y\) through their recurrent structure, effectively unrolling the network over time \cite{7, 14}.

However, despite their suitability for several tasks, RNNs encounter difficulties in handling long sequences due to issues like vanishing or exploding gradients, which hinder their ability to capture long-term dependencies. For instance, in a phrase like ``there are various tourist attractions in the Draa-Tafilalet region, they are diverse and offer a rich experience in cultural discoveries, adventures, and moments of relaxation", RNNs may overlook the connection between ``attractions" and ``they."

To address this limitation, LSTM networks can be employed to enable models to effectively learn and retain both short- and long-term information \cite{15}, making them suitable for tasks such as generating coherent and contextually appropriate responses in chatbots. By leveraging the capabilities of LSTM networks, AI-based chatbots can better manage long sequences and provide more accurate and contextually relevant interactions.

\subsection{Seq2Seq Model}

The Seq2Seq model \cite{14} is a fundamental architecture for dialogue systems and machine translation. It consists of two recurrent neural networks: an encoder and a decoder. The encoder processes the input sequence symbol by symbol, transforming it into a context vector that encapsulates the sequence's intent. Subsequently, the decoder uses this context vector to generate the corresponding output sequence.

In recent years, transformer architecture \cite{16} has emerged as a significant advancement in NLP models. By leveraging attention mechanisms, transformer models excel in capturing comprehensive context when generating or understanding text. This architecture has been instrumental in developing prominent language models such as Generative Pre-trained Transformers (GPT) \cite{17} and Bidirectional Encoder Representations from Transformers (BERT) \cite{18}, both of which have demonstrated exceptional performance in conversational AI tasks \cite{19}. For instance, the advanced capabilities of GPT have facilitated the creation of sophisticated and widely-used chatbots, such as ChatGPT.

\subsection{Uniqueness of our Work}

Unlike existing studies, our work distinguishes itself by focusing specifically on the tourism sector in Draa-Tafilalet, Morocco, leveraging a Seq2Seq model with LSTM and attention mechanisms to achieve a nuanced understanding and generation of contextually relevant responses. While many current chatbots and conversational agents are designed to be general-purpose and often rely on predefined APIs, our approach integrates domain-specific knowledge and cultural context without such dependencies. This not only enhances the accuracy and relevance of the information provided but also offers a cost-effective and flexible solution free from the constraints of commercial NLP services. By addressing the unique needs of the tourism industry in this region, our work provides a more tailored and effective user experience compared to more generic solutions.

\begin{algorithm}[h]
\footnotesize
\caption{\textcolor{black}{Seq2Seq model-based Chatbot with LSTM and attention mechanism.}}
\label{alg:seq2seq}
\begin{algorithmic}[1]
\State \textbf{Input}: Dataset $\mathcal{D} = \{(q_1, a_1), (q_2, a_2), ..., (q_n, a_n)\}$ 
\State \textbf{Output}: Trained Seq2Seq model for chatbot.

\Procedure{Preprocess Data}{}
    \For{each $(q_i, a_i) \in \mathcal{D}$} 
        \State Clean and tokenize $q_i$ and $a_i$ 
        \State Vectorize tokens using GloVe embeddings 
        \State Pad sequences to uniform length $L$ 
    \EndFor
    \State Split data into training set $\mathcal{T}$, validation set $\mathcal{V}$, and test set $\mathcal{E}$ 
\EndProcedure

\Procedure{Train Seq2Seq Model}{}
    \State \textbf{Initialize}: Encoder $Enc$ and decoder $Dec$ with LSTM cells and attention mechanism 
    \For{each epoch $e$ in training} 
        \For{each batch $b \in \mathcal{T}$} 
            \State Encode input sequences $\mathbf{X}_b$ to context vectors $\mathbf{h}_t$ 
            \State Decode context vectors $\mathbf{h}_t$ to output sequences $\hat{\mathbf{Y}}_b$ 
            \State Calculate loss $\mathcal{L} = \sum_{i=1}^{|b|} \text{CrossEntropy}(\hat{y}_i, y_i)$ 
            \State Update model parameters using Adam optimizer to minimize $\mathcal{L}$ 
        \EndFor
        \State Validate model on $\mathcal{V}$ and adjust hyperparameters if necessary 
    \EndFor
\EndProcedure

\Procedure{Evaluate Model}{}
    \For{each $(q_i, a_i) \in \mathcal{E}$} 
        \State Generate predicted answer $\hat{a}_i$ using the trained model 
        \State Compare $\hat{a}_i$ with $a_i$ to calculate accuracy $\mathcal{A} = \frac{1}{|\mathcal{E}|} \sum_{i=1}^{|\mathcal{E}|} \mathbf{1}(\hat{a}_i = a_i)$ 
    \EndFor
    \State Compute overall test accuracy $\mathcal{A}$ 
\EndProcedure

\Procedure{Chatbot Interaction}{}
    \State \textbf{Input}: User query $q$ 
    \State Preprocess and tokenize $q$ 
    \State Generate response $\hat{a}$ using the trained model 
    \State Return $\hat{a}$ as chatbot response 
\EndProcedure

\State \textbf{return} Trained Seq2Seq model
\end{algorithmic}
\end{algorithm}

\section{Proposed Methodology} \label{sec:method}

In this study, our overall methodology comprises five distinct steps. Initially, the dataset undergoes preprocessing procedures. The dataset is then partinioned into three distinct sets, namely training, validation, and test sets. Next, a Seq2Seq model is constructed using encoder-decoder architecture, with LSTM cells and the attention mechanism. Lastly, the Seq2Seq-based model is trained and thoroughly evaluated.

Algorithm \ref{alg:seq2seq} outlines the development of a Seq2Seq model-based chatbot enhanced with LSTM cells and an attention mechanism. The process begins with data preprocessing, including cleaning, tokenizing, vectorizing, and padding the input data. The data is then split into training, validation, and test sets. During the training phase, the model iteratively encodes input sequences and decodes context vectors to generate output sequences, optimizing parameters to minimize loss using the Adam optimizer. The trained model is then evaluated on the test set to calculate its accuracy. Finally, the model is used to generate responses to user queries in real-time, completing the chatbot interaction loop.

\subsection{Data collection}

To prepare the dataset, we adopted an approach based on the Six A Framework for tourist destination analysis \cite{20}: attractions, accessibility, amenities, activities, available packages, and ancillary services. The key features of the datasets are summarized in Table \ref{tab:dataset_summary} and detailed as follows.
\begin{itemize}
    \item \textbf{Attractions}: Information on tourist attractions, including landmarks, historical sites, and natural wonders, was meticulously gathered through extensive research from tourism websites, travel guides, and official destination sources. This effort resulted in 1,432 training examples comprising question-answer (QA) pairs to cover a diverse range of attractions.
    \item \textbf{Amenities}: We identified available amenities and services for tourists, such as accommodations, dining options, and hotels. A total of 338 training examples, each containing a QA pair, were collected to provide comprehensive information about amenities in the Draa-Tafilalet region.
    \item \textbf{Accessibility}: Information on transportation options, routes, and accessibility features was compiled to facilitate tourists' travel plans. This included details about airports, train stations, bus services, and car rentals. We prepared 772 training examples, each with a QA pair, to address various accessibility-related inquiries.
    \item \textbf{Activities}: A list of leisure activities available to tourists, including adventures, cultural experiences, guided tours, and entertainment options, was assembled. This resulted in 420 training examples featuring QA pairs, covering a wide range of activities and experiences offered in the destinations.
    \item \textbf{Available packages}: We collected detailed information about travel deals, including itineraries, pricing, inclusions, and booking details. A total of 226 training examples, each with a QA pair, were collected to provide insights into the available packages and assist users in making informed decisions.
    \item \textbf{Ancillary services}: Ancillary services that complement tourists' experiences, such as tour guides, translators, travel insurance, and local assistance services, were identified. We prepared 512 training examples, each containing a QA pair, to address inquiries related to ancillary services and support tourists' needs during their travels.
\end{itemize}

Overall, the dataset comprises 3,700 utterances, including 1,850 questions and 1,850 responses, ensuring a comprehensive and diverse representation of the tourism-related information for the Draa-Tafilalet region.

\begin{table}[ht]
\centering
\caption{Summary of dataset features for Draa-Tafilalet tourism \cite{DraaTafilalet}.}
\begin{tabular}{@{}lp{3.3cm}c@{}}
\toprule
\textbf{Feature}       & \textbf{Description}                                      & \textbf{Examples} \\ \midrule
Attractions            & Landmarks, historical sites, natural wonders              & 1,432                      \\
Amenities              & Accommodations, dining options, hotels                    & 338                        \\
Accessibility          & Transportation options, routes, accessibility features    & 772                        \\
Activities             & Adventures, cultural experiences, guided tours, entertainment & 420                        \\
Available packages     & Travel deals, itineraries, pricing, inclusions, bookings  & 226                        \\
Ancillary services     & Tour guides, translators, travel insurance, local assistance & 512                        \\ \midrule
\textbf{Total}         &                                                          & \textbf{3,700}             \\ \bottomrule
\end{tabular}
\label{tab:dataset_summary}
\end{table}

\subsection{Data Preprocessing}

Data preprocessing is a critical stage in preparing data for training the Seq2Seq model, involving several meticulous substeps. Initially, a dictionary is constructed to map each conversational utterance to its corresponding identifier within the dataset. Following this, all conversational exchanges (Q/A) are organized into a nested list structure, separating data into distinct question and answer lists. Preprocessing notably includes the removal of uppercase characters, punctuation marks, and special characters from both question and answer lists to ensure data cleanliness.

To maintain uniformity in sequence length, longer utterances are truncated, while shorter ones are padded using a designated token. Additionally, a vocabulary list comprising all unique terms present within the dataset is created. The final preprocessing step involves transforming words in the question-answer lists into vector representations, a process known as vectorization. This transformation can be achieved through either localist representations, such as one-hot encoding, or distributed representations, exemplified by GloVe or Word2Vec. In this study, GloVe embeddings are employed to generate vector representations for the words within our dataset.

\subsection{Data splitting}
The process of data splitting involves partitioning a dataset into three distinct subsets: a training set, a validation set, and a testing set. The training set comprises a substantial portion of the dataset and is utilized exclusively for training the model. Conversely, the validation set is used to fine-tune the model's hyperparameters, facilitating the optimization of its performance. Finally, the testing set is reserved for evaluating the model's efficacy. In the context of our experimentation, the dataset is divided into 98\% for the training set, 1\% for the validation set, and 1\% for the testing set.

\subsection{Model Construction}
\begin{figure}[htbp]
\centering
\centerline{\includegraphics[width=1\linewidth]{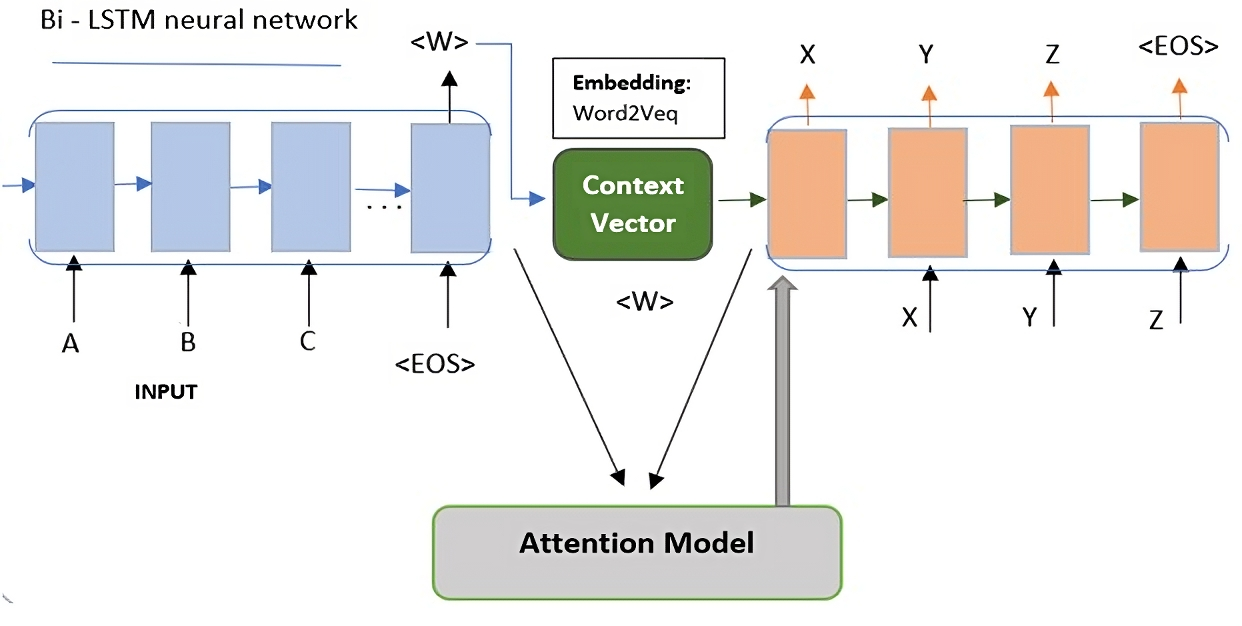}}
\caption{Seq2Seq model architecture with attention mechanism.}
\label{fig:model-architecure}
\end{figure}

We utilized Keras and TensorFlow for the development and training of deep learning models, specifically in constructing Seq2Seq architectures employing encoder-decoder configurations with attention mechanisms using LSTM cells. Fig. \ref{fig:model-architecure} illustrates the architecture of our model, comprising an encoder and a decoder. The encoder module is responsible for processing input sequences, such as English sentences, and transforming them into concise contextual vectors encapsulating salient information. Subsequently, the decoder component leverages these contextual vectors to generate output sequences, serving as coherent responses to user inquiries.

\subsection{Model Training Process}

The model was trained using the preprocessed dataset, with various hyperparameters assigned different values, including the learning rate, number of epochs, batch size, and number of LSTM cells. Categorical cross-entropy was chosen as the loss function, with accuracy serving as the evaluation metric. The Adam optimizer was employed for optimization. Table \ref{tab:hyper-configs} outlines three distinct configurations (C1, C2, and C3) of these hyperparameters used during the model training process.

\begin{table}[htbp]
\caption{Hyperparameter configurations overview.}
\begin{center}
\begin{tabular}{|p{2cm}|p{1.5cm}|p{1.5cm}|p{1.5cm}|}
\hline
\textbf{} & \textbf{C1} & \textbf{C2} & \textbf{C3} \\
\cline{2-4} 
\hline
LSTM cells&	256	&512	&512 \\
\hline
Bidirectional LSTM cells & 512 & 1024 & 1024 \\		
\hline
Batch size &	8	&8	&16\\
\hline
Epochs	 &10	&20	&50\\
\hline
Learning rate&	1e-3	&1e-3	&1e-4\\
\hline
Loss function	& \multicolumn{3}{|c|}{categorical\_crossentropy}\\
\hline
Optimizer &	\multicolumn{3}{|c|}{adam} \\
\hline
Metric	 & \multicolumn{3}{|c|}{accuracy} \\
\hline
\end{tabular}
\label{tab:hyper-configs}
\end{center}
\end{table}

\section{Performance Evaluation and Discussion} \label{sec:discussion}

\begin{figure}[htbp]
\begin{center}
\resizebox{\columnwidth}{!}{
\begin{tikzpicture}
    \begin{axis}[
        ybar,
        symbolic x coords={C 1, C 2, C 3},
        xtick=data,
        ymin=0,
        ymax=100,
        ymajorgrids=true,
        ylabel={Accuracy (\%)},
        xlabel={Configurations},
        legend style={at={(0.5,1.15)},
        anchor=south,legend columns=-1},
        bar width=15pt,
        width=\textwidth,
        height=0.6\textwidth,
        nodes near coords,
        every node near coord/.append style={font=\scriptsize, /pgf/number format/fixed},
        xticklabel style={font=\small, rotate=0},
        enlarge x limits=0.25,
        enlarge y limits=0.1,
        grid style=dashed,
        axis line style=thick,
        title style={font=\bfseries}
    ]
    \addplot[style={fill=blue!30, draw=black}] coordinates {(C 1, 98.72) (C 2, 99.5) (C 3, 99.6)};
    \addplot[style={fill=green!30, draw=black}] coordinates {(C 1, 75.4) (C 2, 98.9) (C 3, 96.3)};
    \addplot[style={fill=red!30, draw=black}] coordinates {(C 1, 72.4) (C 2, 94.1) (C 3, 92.4)};
    \legend{Training accuracy, Validation accuracy, Test accuracy}
    \end{axis}
\end{tikzpicture}
}
\end{center}
\caption{Accuracy of the models over the three configurations.}
\label{fig:accuracy}
\end{figure}
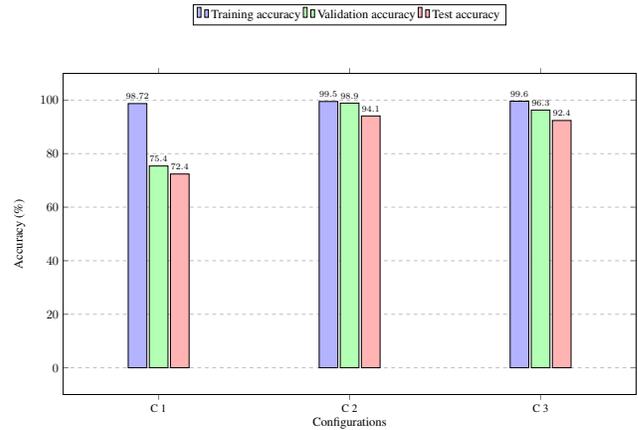

\begin{figure}[htbp]
    \centering
\resizebox{\columnwidth}{!}{
    \begin{tikzpicture}
        \node[draw, fill=black!10, rounded corners, drop shadow, text width=15cm, font=\sffamily, anchor=north west] (box1) at (0, 0) {
            \textbf{Me:} What are some famous natural attractions in Draa-Tafilalet? \\
            \textbf{chatbot:} Famous natural attractions in draa-tafilalet include the draa valley, todra gorge, and tafilalet. \textless{}UNK\textgreater{} \textless{}PAD\textgreater{} \textless{}PAD\textgreater{} \textless{}PAD\textgreater{}
        };

        \node[draw, fill=white, rounded corners, drop shadow, text width=15cm, font=\sffamily, anchor=north west, below=0.5cm of box1] (box2) {
            \textbf{Me:} What activities can I enjoy in Todra Gorge? \\
            \textbf{chatbot:} Todra gorge visitors can enjoy activities such as hiking along the canyon, rock climbing on its towering cliffs, and admiring the stunning natural scenery.
        };

        \node[draw, fill=black!10, rounded corners, drop shadow, text width=15cm, font=\sffamily, anchor=north west, below=0.5cm of box2] (box3) {
            \textbf{Me:} Can you swim in the natural pools of Todra Gorge? \\
            \textbf{chatbot:} In the natural pools of todra gorge, swimming is not recommended due to strong currents and rocky terrain, but you can enjoy the views and take a dip in safer areas.
        };

        \node[draw, fill=white, rounded corners, drop shadow, text width=15cm, font=\sffamily, anchor=north west, below=0.5cm of box3] (box4) {
            \textbf{Me:} What is the history behind Tafilalet Oasis? \\
            \textbf{chatbot:} Tafilalet oasis has a rich history dating back thousands of years, serving as a vital oasis on the trans-Saharan trade routes and a center of agriculture.
        };

        \node[draw, fill=black!10, rounded corners, drop shadow, text width=15cm, font=\sffamily, anchor=north west, below=0.5cm of box4] (box5) {
            \textbf{Me:} What role did kasbahs play in the region's history? \\
            \textbf{chatbot:} Kasbahs served as fortified residences for local chieftains and nobility, as well as strategic strongholds for defense against invaders, contributing to the region's political and social history.
        };

        \node[draw, fill=white, rounded corners, drop shadow, text width=15cm, font=\sffamily, anchor=north west, below=0.5cm of box5] (box6) {
            \textbf{Me:} What are the main features of Jebel Saghro? \\
            \textbf{chatbot:} Jebel saghro is known for its rugged terrain, rocky peaks, and deep gorges, offering adventurous opportunities for hiking, climbing, and exploring.
        };
    \end{tikzpicture}
}
    \caption{Sample Chatbot conversation with DraaBot discussing famous attractions in Draa-Tafilalet \cite{DraaTafilalet}.}
    \label{fig:chatbot_conversation}
\end{figure}

In this experimental investigation, optimal performance was achieved with a specific set of hyperparameters. This configuration included a learning rate of 1e-3, 20 epochs, a batch size of 8, and 512 LSTM cells. Following the training phase, the model exhibited a training accuracy of approximately 99.58\%, with a corresponding validation accuracy of roughly 98.03\%, as depicted in Fig. \ref{fig:accuracy}. Evaluation on the test dataset yielded an accuracy of 94.12\%. The experimental results are summarized in Table \ref{tab:hyper-configs}.

Furthermore, to illustrate the practical application of our model, Fig. \ref{fig:chatbot_conversation} shows a sample conversation with our proposed chatbot. This interaction highlights the chatbot's ability to generate relevant and coherent responses within the tourism domain, demonstrating the effectiveness of the trained Seq2Seq model with LSTM and attention mechanisms. Next, we delve into the implications of our results obtained from training the chatbot models with different configurations. The configurations (named C1, C2, and C3), present varying levels of performance across training, validation, and test sets.

In Configuration 1, we observed a training accuracy of 98.72\%, with validation and test accuracies of 75.43\% and 72.43\%, respectively. Despite achieving a relatively high training accuracy, the model's performance on unseen data, as reflected in the validation and test accuracies, was notably lower. This suggests a potential issue with overfitting, where the model may have memorized the training data too closely, resulting in reduced generalization to new inputs.

Conversely, Configuration 2 demonstrated significantly improved performance across all metrics, with a training accuracy of 99.58\%, validation accuracy of 98.98\%, and test accuracy of 94.12\%. This configuration achieved remarkable consistency between the training and validation accuracies, indicating robust generalization capabilities. The high test accuracy further corroborates the model's effectiveness in accurately responding to user queries.

In Configuration 3, while the training accuracy remained high at 99.63\%, we observed a slight decrease in both validation and test accuracies, which stood at 96.31\% and 92.43\%, respectively. This discrepancy between training and validation/test accuracies suggests a potential issue with model complexity or insufficient regularization, leading to suboptimal performance on unseen data.

Overall, our results underscore the importance of carefully tuning hyperparameters and model architectures to achieve optimal performance in chatbot. Configuration C2 emerges as the most promising configuration, exhibiting strong generalization capabilities across our dataset for tourism context.

\subsection{Complexity Analysis}

The proposed methodology utilizes a Seq2Seq model with LSTM cells and an attention mechanism. Below is an advanced complexity analysis for each step of the methodology.

\subsubsection{Data preprocessing complexity}
\begin{itemize}
    \item \textbf{Cleaning, Tokenizing, and Padding:}
    Let $n$ be the total number of utterances and $L$ be the maximum sequence length.
    \begin{itemize}
        \item \textit{Tokenization:} Tokenizing each utterance takes $O(L)$, resulting in an overall time complexity of $O(n \times L)$.
        \item \textit{Vectorization with GloVe Embeddings:} Assuming the vocabulary size is $V$, the lookup for each token is $O(1)$. Therefore, the overall complexity for vectorization is $O(n \times L)$.
        \item \textit{Padding Sequences:} Padding requires checking each sequence to extend to the maximum length, resulting in $O(n \times L)$.
    \end{itemize}
\end{itemize}

Overall, data preprocessing has a complexity of $O(n \times L)$.

\subsubsection{Model construction complexity}
\begin{itemize}
    \item \textbf{Encoder-Decoder Architecture with Attention Mechanism:}
    \begin{itemize}
        \item \textit{LSTM Encoder:} The LSTM encoder processes the input sequence in a sequential manner. For an input of length $L$ and hidden state dimension $h$, the time complexity per sequence is $O(L \times h^2)$ due to the recurrent computation for each time step involving matrix multiplication.
        \item \textit{Attention Mechanism:} The attention mechanism involves computing a similarity score for each encoder hidden state with the decoder's current hidden state. This step takes $O(L \times h)$, where $L$ is the length of the input sequence.
        \item \textit{LSTM Decoder with Attention:} Each output token requires attention over all encoder states, which involves computing the attention weights and generating the context vector. The overall complexity per output token is $O(L \times h)$, and for the full output sequence of length $L'$, it is $O(L \times L' \times h)$.
    \end{itemize}
\end{itemize}

Thus, the overall complexity for the encoder-decoder with attention mechanism is $O(L \times h^2) + O(L \times L' \times h)$.

\subsubsection{Model training complexity}
\begin{itemize}
    \item \textbf{Forward and Backward Propagation:}
    \begin{itemize}
        \item During training, the forward pass involves computing the outputs for both the encoder and decoder, resulting in a time complexity of $O(n \times (L \times h^2 + L \times L' \times h))$.
        \item Backpropagation involves calculating gradients for each weight parameter in the LSTM and attention mechanism. The complexity for backpropagation is approximately the same as the forward pass, leading to an overall complexity of $O(n \times (L \times h^2 + L \times L' \times h))$.
    \end{itemize}
    \item \textbf{Optimizer Complexity (Adam):} The Adam optimizer maintains two running averages for each parameter, which results in a complexity of $O(P)$, where $P$ is the total number of parameters in the model. Since $P$ depends on the number of LSTM units and the input/output dimensions, this step adds a complexity of $O(P)$ per update.
\end{itemize}

The overall complexity for training with $E$ epochs and $B$ batches per epoch is:
\[
O(E \times B \times n \times (L \times h^2 + L \times L' \times h) + E \times B \times P)
\]

\subsubsection{Complexity summary}
\begin{itemize}
    \item \textbf{Data preprocessing:} $O(n \times L)$
    \item \textbf{Model construction (encoder-decoder with attention):} $O(L \times h^2) + O(L \times L' \times h)$
    \item \textbf{Model training:} $O(E \times B \times n \times (L \times h^2 + L \times L' \times h) + E \times B \times P)$
\end{itemize}

\section{Conclusion} \label{sec:conclusion}
This paper presented a Seq2Seq model-based chatbot solution, which leverages LSTM cells and attention mechanisms, specifically tailored for the tourism sector in Draa-Tafilalet, Morocco. Through rigorous experimentation and evaluation, the proposed chatbot demonstrated high accuracy and effective interaction capabilities, outperforming general-purpose chatbots by integrating domain-specific knowledge and cultural context. This work not only addresses the limitations of existing chatbots, such as dependency on predefined APIs and high costs, but also provides a flexible and cost-effective solution.

Future research will explore advanced attention mechanisms, multi-turn dialogue capabilities, and context awareness to further enhance the chatbot's performance and user engagement in the tourism domain. The success of this chatbot solution underscores the potential of specialized AI solutions in delivering superior user experiences in niche sectors.

\section*{Acknowledgment}
This work was supported by the Ministry of Higher Education, Scientific Research and Innovation, the Digital Development Agency (DDA), and the CNRST of Morocco (Alkhawarizmi/2020/32).

\bibliography{aaai25}

\begin{thebibliography}{25}
\providecommand{\natexlab}[1]{#1}

\bibitem[{Adamopoulou and Moussiades(2020)}]{adamopoulou2020chatbots}
Adamopoulou, E.; and Moussiades, L. 2020.
\newblock Chatbots: History, technology, and applications.
\newblock \emph{Machine Learning with applications}, 2: 100006.

\bibitem[{Benaddi et~al.(2023)Benaddi, Ouaddi, Khriss, and Ouchao}]{2}
Benaddi, L.; Ouaddi, C.; Khriss, I.; and Ouchao, B. 2023.
\newblock Analysis of Tools for the Development of Conversational Agents.
\newblock In \emph{Computer Sciences and Mathematics Forum}, volume~6, 5. MDPI.

\bibitem[{Brown et~al.(2020)}]{17}
Brown, T.; et~al. 2020.
\newblock Language models are few-shot learners.
\newblock \emph{Advances in neural information processing systems}, 33: 1877--1901.

\bibitem[{Buhalis(2000)}]{20}
Buhalis, D. 2000.
\newblock Marketing the competitive destination of the future.
\newblock \emph{Tourism management}, 21(1): 97--116.

\bibitem[{Caldarini, Jaf, and McGarry(2022)}]{6}
Caldarini, G.; Jaf, S.; and McGarry, K. 2022.
\newblock A literature survey of recent advances in chatbots.
\newblock \emph{Information}, 13(1): 41.

\bibitem[{Casheekar et~al.(2024)Casheekar, Lahiri, Rath, Prabhakar, and Srinivasan}]{casheekar2024contemporary}
Casheekar, A.; Lahiri, A.; Rath, K.; Prabhakar, K.~S.; and Srinivasan, K. 2024.
\newblock A contemporary review on chatbots, AI-powered virtual conversational agents, ChatGPT: Applications, open challenges and future research directions.
\newblock \emph{Computer Science Review}, 52: 100632.

\bibitem[{Devlin et~al.(2018)Devlin, Chang, Lee, and Toutanova}]{18}
Devlin, J.; Chang, M.-W.; Lee, K.; and Toutanova, K. 2018.
\newblock Bert: Pre-training of deep bidirectional transformers for language understanding.
\newblock \emph{arXiv preprint arXiv:1810.04805}.

\bibitem[{Hochreiter and Schmidhuber(1997)}]{15}
Hochreiter, S.; and Schmidhuber, J. 1997.
\newblock Long short-term memory.
\newblock \emph{Neural computation}, 9(8): 1735--1780.

\bibitem[{Lebeuf, Storey, and Zagalsky(2017)}]{1}
Lebeuf, C.; Storey, M.-A.; and Zagalsky, A. 2017.
\newblock Software bots.
\newblock \emph{IEEE Software}, 35(1): 18--23.

\bibitem[{Limna et~al.(2023)Limna, Kraiwanit, Jangjarat, Klayklung, and Chocksathaporn}]{limna2023use}
Limna, P.; Kraiwanit, T.; Jangjarat, K.; Klayklung, P.; and Chocksathaporn, P. 2023.
\newblock The use of ChatGPT in the digital era: Perspectives on chatbot implementation.
\newblock \emph{Journal of Applied Learning and Teaching}, 6(1): 64--74.

\bibitem[{Ouaddi, Benaddi, and Jakimi(2024)}]{10}
Ouaddi, C.; Benaddi, L.; and Jakimi, A. 2024.
\newblock Architecture, tools, and dsls for developing conversational agents: An overview.
\newblock \emph{Procedia Computer Science}, 231: 293--298.

\bibitem[{Ouaddi et~al.(2023{\natexlab{a}})Ouaddi, Benaddi, Khriss, and Jakimi}]{7}
Ouaddi, C.; Benaddi, L.; Khriss, I.; and Jakimi, A. 2023{\natexlab{a}}.
\newblock Developing Conversational Agent Using Deep Learning Techniques.
\newblock In \emph{Computer Sciences and Mathematics Forum}, volume~6, 3. MDPI.

\bibitem[{Ouaddi et~al.(2023{\natexlab{b}})Ouaddi, Benaddi, Souha, and Jakimi}]{13}
Ouaddi, C.; Benaddi, L.; Souha, A.; and Jakimi, A. 2023{\natexlab{b}}.
\newblock Towards a Unified UML-based Approach to Modeling a Conversation Scenario for Chatbot.
\newblock In Springer, ed., \emph{International Conference on Advanced Technologies for Humanity}.

\bibitem[{Ouaddi et~al.(2024)Ouaddi, Benaddi, Souha, and Jakimi}]{12}
Ouaddi, C.; Benaddi, L.; Souha, A.; and Jakimi, A. 2024.
\newblock Towards a Metamodel for Proactive Chatbots.
\newblock In \emph{2024 ASU International Conference in Emerging Technologies for Sustainability and Intelligent Systems (ICETSIS)}, 1129--1133. IEEE.

\bibitem[{Pérez-Soler, Guerra, and de~Lara(2020)}]{11}
Pérez-Soler, S.; Guerra, E.; and de~Lara, J. 2020.
\newblock Model-Driven Chatbot Development.
\newblock In Dobbie, G.; Frank, U.; Kappel, G.; Liddle, S.~W.; and Mayr, H.~C., eds., \emph{Conceptual Modeling}, 207--222. Cham: Springer International Publishing.

\bibitem[{Revang et~al.(2018)Revang, Baker, Manusama, and Mullen}]{8}
Revang, M.; Baker, V.; Manusama, B.; and Mullen, A. 2018.
\newblock Market guide for conversational platforms.
\newblock Technical report, Gartner.

\bibitem[{Revang, Mullen, and Elliot(2022)}]{9}
Revang, M.; Mullen, A.; and Elliot, B. 2022.
\newblock Magic quadrant for enterprise conversational ai platforms.
\newblock Technical report, Gartner Reports.

\bibitem[{Sarikaya(2017)}]{4}
Sarikaya, R. 2017.
\newblock The technology behind personal digital assistants: An overview of the system architecture and key components.
\newblock \emph{IEEE Signal Processing Magazine}, 34(1): 67--81.

\bibitem[{Setiawan, Utami, and Hartanto(2021)}]{3}
Setiawan, B.~A.; Utami, E.; and Hartanto, A.~D. 2021.
\newblock Banjarese chatbot using seq2seq model.
\newblock In \emph{2021 4th International Conference on Information and Communications Technology (ICOIACT)}, 233--238. IEEE.

\bibitem[{Sojasingarayar(2020)}]{14}
Sojasingarayar, A. 2020.
\newblock Seq2seq ai chatbot with attention mechanism.
\newblock \emph{arXiv preprint arXiv:2006.02767}.

\bibitem[{Souha et~al.(2023)Souha, Ouaddi, Benaddi, and Jakimi}]{19}
Souha, A.; Ouaddi, C.; Benaddi, L.; and Jakimi, A. 2023.
\newblock Pre-Trained Models for Intent Classification in Chatbot: Comparative Study and Critical Analysis.
\newblock In \emph{2023 6th International Conference on Advanced Communication Technologies and Networking (CommNet)}, 1--6. IEEE.

\bibitem[{Unknown()}]{DraaTafilalet}
Unknown. ????
\newblock Draa-Tafilalet in Morocco.
\newblock \url{https://www.visitmorocco.com/en/regions/dr%C3%A2a-tafilalet}.
\newblock Accessed: 2024-05-22.

\bibitem[{Vaswani et~al.(2017)}]{16}
Vaswani, A.; et~al. 2017.
\newblock Attention is all you need.
\newblock In \emph{Proceedings of the 31st International Conference on Neural Information Processing Systems}. Long Beach, California, USA.

\bibitem[{Weizenbaum(1966)}]{5}
Weizenbaum, J. 1966.
\newblock ELIZA—a computer program for the study of natural language communication between man and machine.
\newblock \emph{Communications of the ACM}, 9(1): 36--45.

\bibitem[{Zhai and Wibowo(2022)}]{zhai2022systematic}
Zhai, C.; and Wibowo, S. 2022.
\newblock A systematic review on cross-culture, humor and empathy dimensions in conversational chatbots: the case of second language acquisition.
\newblock \emph{Heliyon}, 8(12).

\end{thebibliography}

\end{document}